\documentclass[journal,twocolumn,final]{IEEEtran}
\usepackage{graphicx,times,amsmath,amssymb,cite,subfigure,flushend,url}
\usepackage[lined,ruled,commentsnumbered]{algorithm2e}
\IEEEoverridecommandlockouts \allowdisplaybreaks 

\begin{document}

\title{Spatial Filtering for EEG-Based Regression Problems in Brain-Computer Interface (BCI)}

\author{\IEEEauthorblockN{Dongrui Wu\IEEEauthorrefmark{1}, \textit{Senior Member, IEEE}, Jung-Tai King\IEEEauthorrefmark{2}, Chun-Hsiang Chuang\IEEEauthorrefmark{3}\IEEEauthorrefmark{2}, \\ Chin-Teng Lin\IEEEauthorrefmark{3}\IEEEauthorrefmark{2}, \textit{Fellow, IEEE}}, Tzyy-Ping Jung\IEEEauthorrefmark{4}\IEEEauthorrefmark{5}, \textit{Fellow, IEEE} \\
\IEEEauthorblockA{\IEEEauthorrefmark{1}DataNova, NY USA}\\
\IEEEauthorblockA{\IEEEauthorrefmark{2}Brain Research Center, National Chiao-Tung University, Hsinchu, Taiwan}\\
\IEEEauthorblockA{\IEEEauthorrefmark{3}Faculty of Engineering and Information Technology, University of Technology, Sydney, Australia}\\
\IEEEauthorblockA{\IEEEauthorrefmark{4}Swartz Center for Computational Neuroscience, Institute for Neural Computation, University of California San Diego, La Jolla, CA}\\
\IEEEauthorblockA{\IEEEauthorrefmark{5}Center for Advanced Neurological Engineering, Institute of Engineering in
Medicine, University of California San Diego, La Jolla, CA}\\
E-mail: drwu09@gmail.com, jtchin2@gmail.com, cch.chuang@gmail.com,\\ Chin-Teng.Lin@uts.edu.au, jung@sccn.ucsd.edu}
\maketitle

\begin{abstract}
Electroencephalogram (EEG) signals are frequently used in brain-computer interfaces (BCIs), but they are easily contaminated by artifacts and noises, so preprocessing must be done before they are fed into a machine learning algorithm for classification or regression. Spatial filters have been widely used to increase the signal-to-noise ratio of EEG for BCI classification problems, but their applications in BCI regression problems have been very limited. This paper proposes two common spatial pattern (CSP) filters for EEG-based regression problems in BCI, which are extended from the CSP filter for classification, by making use of fuzzy sets. Experimental results on EEG-based response speed estimation from a large-scale study, which collected 143 sessions of sustained-attention psychomotor vigilance task data from 17 subjects during a 5-month period, demonstrate that the two proposed spatial filters can significantly increase the EEG signal quality. When used in LASSO and $k$-nearest neighbors regression for user response speed estimation, the spatial filters can reduce the root mean square estimation error by $10.02-19.77\%$, and at the same time increase the correlation to the true response speed by $19.39-86.47\%$.
\end{abstract}

\begin{IEEEkeywords}
Brain-computer interface, common spatial pattern, EEG, fuzzy sets, psychomotor vigilance task, response speed estimation, spatial filtering
\end{IEEEkeywords}

\section{Introduction}

Electroencephalogram (EEG) signals are the most widely used input for brain-computer interfaces (BCIs) \cite{Lance2012,Erp2012,Liao2012,Makeig2012,Tan2010,Nicolas-Alonso2012}, mainly due to the convenience to obtain them, compared with magnetoencephalography (MEG) \cite{Mellinger2007}, functional magnetic resonance imaging (fMRI) \cite{Sitaram2007}, functional near-infrared spectroscopy (fNIRS) \cite{Naseer2015}, and invasive signals like electrocorticography (ECoG) \cite{Pei2011} and intracortical neural recordings \cite{Maynard1997}. However, EEG signals are often contaminated by ocular, muscular, and cardiac artifacts and various noises (power-line, changes in electrode impedances, etc) \cite{Uriguen2015,Bigdely-Shamlo2015,Nicolas-Alonso2012}. Usually some preprocessing, either manually or automatically \cite{Bigdely-Shamlo2015,Nicolas-Alonso2012}, is needed to remove the artifacts, and then temporal and spatial filters are applied to further improve the EEG signal quality before feeding it into a classification or regression algorithm. The most commonly used temporal filters are band-pass filters and notch filters (at 50 or 60 Hz power-line frequency).

This paper focuses on spatial filtering for improving the EEG signal quality. Many such approaches have been proposed in the literature \cite{Vigario2000,Hyvarinen2000,Rivet2009,Rivet2013,Roy2015,Hotelling1936,Blankertz2008,Ramoser2000,Barachant2014b}. However, almost all of them focus primarily on EEG classification problems in BCI, whereas EEG regression problems have been largely overlooked. Nevertheless, the latter is also very important in BCI. One example is driver drowsiness (or alertness) estimation from EEG signals, which has been extensively studied in our previous research \cite{drwuaBCI2015,Lin2008,Lin2005d,drwuTFS2016,drwuEBMAL2016,Lin2006,Wei2015}. This is a very important problem because drowsy driving is among the most important causes of road crashes, following only to alcohol, speeding, and inattention \cite{Sagberg2004}. According to the National Highway Traffic Safety Administration \cite{NHTSA2011}, 2.5\% of fatal motor vehicle crashes (on average 886/year in the U.S.) and 2.5\% of fatalities (on average 1,004/year in the U.S.) between 2005 and 2009 involved drowsy driving.

This paper proposes two spatial filters for EEG-based regression problems in BCI. We also validate their performance in response speed (RS) estimation from EEG signals measured in a large-scale sustained-attention psychomotor vigilance task (PVT) \cite{Kerick2016}, which collected 143 sessions of data from 17 subjects in a 5-month period.

The remainder of this paper is organized as follows: Section~\ref{sect:Review} reviews the state-of-the-art spatial filters for EEG-based classification problems in BCI. Section~\ref{sect:Filter} introduces our proposed spatial filters for supervised BCI regression problems. Section~\ref{sect:exp} describes the experimental setup, RS and EEG data preprocessing techniques, and the procedure to evaluate the performances of different spatial filters. Section~\ref{sect:results} presents the results of the comparative studies and parameter sensitivity analysis for the proposed spatial filter.
Section~\ref{sect:discussions} discusses the limitations of the proposed approaches and outlines several future research directions. Finally, Section~\ref{sect:conclusions} draws conclusions.

\section{Spatial Filters for EEG Classification in BCI} \label{sect:Review}

Many spatial filters have been proposed for EEG classification in BCI. The most basic ones include common average reference (CAR) \cite{Teplan2002}, Laplacian filters \cite{Lagerlund1997}, and principal component analysis \cite{Jolliffe2002}. Some of the more recent and also more sophisticated ones are:
\begin{enumerate}
\item \emph{Independent Component Analysis (ICA)} \cite{Delorme2004,Vigario2000,Hyvarinen2000}, which decomposes a multivariate signal into independent non-Gaussian signals. ICA has been widely used in the EEG research community to detect and remove stereotyped eye, muscle, and line noise artifacts \cite{Jung2000,Uriguen2015,Lin2005d}.

    Generally ICA works on an unepoched long block of EEG data, instead of epoched short EEG trials. Let the unepoched EEG data be $\mathbf{X}\in \mathbb{R}^{C\times T}$, where $C$ is the number of EEG channels, and $T$ is the number of time samples. ICA assumes that $\mathbf{X}$ is the linear combination of $c$ independent sources, i.e.,
    $ \mathbf{X}=\mathbf{AS}$, where $\mathbf{A}\in\mathbb{R}^{C\times c}$ is the \emph{mixing matrix}, and the source signals, which are the rows of $\mathbf{S}\in\mathbb{R}^{c\times T}$, are supposed to be stationary, independent, and non-Gaussian. ICA can use various different principles \cite{Uriguen2015,Hyvarinen2000,Vigario2000,Delorme2004} to estimate both unknown $\mathbf{A}$ and unknown $\mathbf{S}$ simultaneously from $\mathbf{X}$. Once $\mathbf{S}$ is obtained, cleaner and more representative features may be extracted from it than from the original $\mathbf{X}$ \cite{Lin2005d}.

\item \emph{xDAWN algorithm} \cite{Rivet2009,Rivet2011,Rivet2013}, which is often used to increase the signal to signal-plus-noise ratio in P300-based BCIs.

    Like ICA, xDAWN also works on the unepoched long block of EEG data $\mathbf{X}\in \mathbb{R}^{C\times T}$. It assumes that $\mathbf{X}=\mathbf{PD}^T+\mathbf{N}$, where $\mathbf{P}\in \mathbb{R}^{C\times S}$ represents the P300 signal in an EEG epoch, and $\mathbf{D}\in\mathbb{R}^{T\times S}$ is a Toeplitz matrix whose first column is defined as:
    \begin{align}
    \mathbf{D}_{\tau_{k},1}=\left\{\begin{array}{ll}
                                     1, & \tau_k \mathrm{\ is\ the\ onset\ of\ the\ }\\
                                     & k\mathrm{th\ target\ stimulus} \\
                                     0, & \mathrm{otherwise}
                                   \end{array}\right.
\end{align}
and $\mathbf{N}\in\mathbb{R}^{C\times T}$ represents the ongoing background brain activity as well as the artifacts and noises. xDAWN then designs a spatial filtering matrix $\mathbf{W}^*\in\mathbb{R}^{C\times F}$, where $F$ is the number of spatial filters, to maximize the signal to signal-plus-noise ratio, i.e.,
\begin{align}
\mathbf{W}^*=\arg\max\limits_{\mathbf{W}}\frac{\mathrm{Tr}(\mathbf{W}^T\mathbf{PD}^T\mathbf{DP}^T\mathbf{W})}
{\mathrm{Tr}(\mathbf{W}^T\mathbf{X} \mathbf{X}^T\mathbf{W})} \label{eq:xDAWN}
\end{align}
where $\mathrm{Tr}(\cdot)$ is the trace of a matrix. (\ref{eq:xDAWN}) is a generalized Rayleigh quotient \cite{Golub1996}, and its solution $\mathbf{W}^*$ is the concatenation of the $F$ eigenvectors associated with the $F$ largest eigenvalues of the matrix $(\mathbf{X}\mathbf{X}^T)^{-1}\mathbf{PD}^T\mathbf{DP}^T$.

The spatially filtered trial for $\mathbf{X}_n$ is then computed as:
\begin{align}
\mathbf{X}_n'={\mathbf{W}^*}^T\mathbf{X}_n,\quad n=1,...,N. \label{eq:Xi}
\end{align}

\item \emph{Canonical Correlation Analysis (CCA)} \cite{Roy2015,Hotelling1936}, which finds linear
transformations to maximize the correlations between two datasets. It has been used to improve BCI performance in code-modulated visual evoked potentials \cite{Bin2011}, steady-state visual evoked potentials \cite{Bin2009a}, and event-related potentials like P300 and error-related potentials \cite{Spuler2014}.

Unlike ICA and xDAWN, CCA works on epoched EEG trials. Consider a binary classification problem, with $N_1$ training examples in Class 1 and $N_2$ training examples in Class 2. Let $(\mathbf{X}_n,y_n)$ be the $n$th training example, where $\mathbf{X}_n\in\mathbb{R}^{C\times S}$ ($C$ is the number of channels, and $S$ is the number of time samples in each trial), and $y_n\in\{1,2\}$. Let $\bar{\mathbf{X}}_k\in\mathbb{R}^{C\times S}$ be the average of $\mathbf{X}_n$ in Class $k$ $(k=1,2)$. We then construct $\tilde{\mathbf{X}}=[\tilde{\mathbf{X}}_1 \ \tilde{\mathbf{X}}_2]$ and $\tilde{\mathbf{Z}}=[\tilde{\mathbf{Z}}_1\ \tilde{\mathbf{Z}}_2]$, where $\tilde{\mathbf{X}}_k$ is the concatenation of all $N_k$ $\mathbf{X}_n$ in Class $k$, and $\tilde{\mathbf{Z}}_k$ is the concatenation of $N_k$ $\bar{\mathbf{X}}_k$. CCA first finds two vector filters $\mathbf{w}_{\tilde{\mathbf{X}}}$ and $\mathbf{w}_{\tilde{\mathbf{Z}}}$ such that the correlation between $\mathbf{w}_{\tilde{\mathbf{X}}}^T\tilde{\mathbf{X}}$ and $\mathbf{w}_{\tilde{\mathbf{Z}}}^T\tilde{\mathbf{Z}}$ is maximized. $\mathbf{w}_{\tilde{\mathbf{X}}}^T\mathbf{X}$ and $\mathbf{w}_{\tilde{\mathbf{Z}}}^T\tilde{\mathbf{Z}}$ are called the first pair of canonical variables. CCA then finds the second pair of canonical variables in a similar way, subject to the constraint that they are uncorrelated with the first pair of canonical variables. This procedure can be continued up to $C$ times.

Finally, the spatial filtering matrix is the concatenation of all $\mathbf{w}_{\tilde{\mathbf{X}}}$, which can be applied to each $\mathbf{X}_n$ to increase its SNR.

\item \emph{Common Spatial Patterns (CSP)} \cite{Blankertz2008,Ramoser2000}, which is a supervised technique frequently used to enhance the binary classification performance of EEG data. The basic idea is to separate the EEG signal into additive subcomponents which have maximum differences in variance between the two classes. In the following we introduce the one-versus-the-rest (OVR) CSP \cite{Dornhege2004}, which extends the traditional CSP from binary classification to $K$ classes.

    Like CCA, OVR CSP also works on epoched EEG trials. Let $(\mathbf{X}_n, y_n)$ be the $n$th training example, as defined above. Assume the mean of $\mathbf{X}_n$ has been removed, e.g., by high-pass or band-pass filtering. Then, for Class $k$, OVR CSP finds a spatial filter matrix $\mathbf{W}_k^*\in\mathbb{R}^{C\times F}$, where $F$ is the number of spatial filters, to maximize the variance difference between Class $k$ and the rest:
    \begin{align}
    \mathbf{W}_k^*
    =\arg\max\limits_{\mathbf{W}}\frac{\mathrm{Tr}(\mathbf{W}^T\bar{\mathbf{\Sigma}}_k\mathbf{W})}
    {\mathrm{Tr}[\mathbf{W}^T(\sum_{i\neq k}\bar{\mathbf{\Sigma}}_i)\mathbf{W}]} \label{eq:CSP}
    \end{align}
    where $\bar{\mathbf{\Sigma}}_k$ is the mean covariance matrix of trials in Class $k$. (\ref{eq:CSP}) is also a generalized Rayleigh quotient \cite{Golub1996}, and the solution $\mathbf{W}_k^*$ is the concatenation of the $F$ eigenvectors associated with the $F$ largest eigenvalues of the matrix $(\sum_{i\neq k}\bar{\mathbf{\Sigma}}_i)^{-1}\bar{\mathbf{\Sigma}}_k$.

    Finally, we concatenate the $K$ individual OVR CSP spatial filters to obtain the complete filter:
    \begin{align}
    \mathbf{W}^*=[\mathbf{W}_1^*,\, ...\, \mathbf{W}_K^*]\in\mathbb{R}^{C\times KF}  \label{eq:W}
    \end{align}
    and compute the spatially filtered trial for $\mathbf{X}_n$ by (\ref{eq:Xi}).
\end{enumerate}

\section{Spatial Filters for Supervised BCI Regression Problems} \label{sect:Filter}

In this section we propose two common spatial pattern for regression (CSPR) filters, which extend the multi-class CSP filters from classification to regression by making use of fuzzy sets \cite{Zadeh1965}, as we have done in \cite{drwuTFS2016}.

First, a brief introduction of fuzzy sets is given below.

\subsection{Fuzzy Sets}

\begin{figure}[htpb]
\centering \includegraphics[width=6cm,clip]{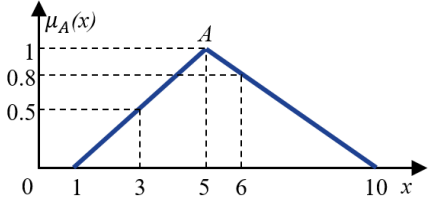} \caption{An examples of a fuzzy set. } \label{fig:T1FS}
\end{figure}

A fuzzy set $A$ is comprised of a \emph{universe of discourse} $D_A$ of real numbers together with a \emph{membership function} $\mu_A:\ D_A \to [0,1]$, i.e.,
\begin{align}
A=\int_{D_A}\mu_A(x)/x
\end{align}
Here $\int$ denotes the collection of all points $x\in D_A$ with associated \emph{membership degree} $\mu_A(x)$. An example of a fuzzy set is shown in Fig.~\ref{fig:T1FS}. The membership degrees are $\mu_A(1)=0$, $\mu_A(3)=0.5$, $\mu_A(5)=1$, $\mu_A(6)=0.8$, and $\mu_A(10)=0$. Observe that this is different from traditional (binary) sets, where each element can only belong to a set completely (i.e., with membership degree 1), or does not belong to it at all (i.e., with membership degree 0); there is nothing in between (i.e., with membership degree 0.5). Fuzzy sets are frequently used in modeling concepts in natural language \cite{Klir1995,Wang1997,Ragin2000}, which may not have clear boundaries.

\subsection{CSPR-OVR}

Let $\mathbf{X}_n\in \mathbb{R}^{C\times S}$ ($n=1,...,N$) be the $n$th EEG trial, where $C$ is the number of channels and $S$ is the number of time samples in each trial. We assume that the mean of each channel measurement has been removed, which is usually performed by band-pass filtering. Let $y_n\in\{1,...,K\}$ be the RS of $\mathbf{X}_n$.

\begin{figure}[htpb]
\centering \includegraphics[width=8cm,clip]{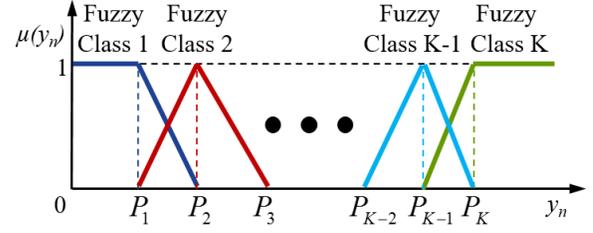} \caption{The $K$ fuzzy classes for $y_n$, when triangular fuzzy sets are used.} \label{fig:FSs}
\end{figure}

With the help of fuzzy sets, we can define ``fuzzy" classes to connect regression problems and classification problems. Assume $K$ fuzzy classes are used. First, we partition the interval $[0, 100]$ into $K+1$ equal intervals, and denote the partition points as $\{p_k\}_{k=1,...,K}$. It is easy to obtain that
\begin{align}
p_k=\frac{100\cdot k}{K+1},\qquad k=1,...,K \label{eq:pk}
\end{align}
For each $p_k$, we then find the corresponding $p_k$ percentile value of all training $y_n$ and denote it as $P_k$. Next we define $K$ fuzzy classes from them, as shown in Fig.~\ref{fig:FSs}. In this way, we can ``classify" the training $y_n$ into $K$ fuzzy classes, corresponding to the $K$ crisp classes in the CSP for classification. However, note that in the CSP for classification a $y_n$ belongs to a crisp class either completely or not at all. For a fuzzy class here, a $y_n$ can belong to it at a membership degree in $[0, 1]$.

Next, for each fuzzy class, we compute its mean EEG trial as:
\begin{align}
\bar{\mathbf{X}}_k=\frac{\sum_{n=1}^N \mu_k(y_n)\mathbf{X}_n}{\sum_{n=1}^N \mu_k(y_n)}, \qquad k=1,...,K\label{eq:fP0}
\end{align}
where $\mu_k(y_n)$ is the membership degree of $y_n$ in Fuzzy Class $k$. Substituting (\ref{eq:fP0}) into (\ref{eq:CSP}), we can solve for the spatial filtering matrix $\mathbf{W}_k^*$ for Fuzzy Class $k$. Essentially, this $\mathbf{W}_k^*$ makes those $\mathbf{X}_n$ in Fuzzy Class $k$ different from those not in Fuzzy Class $k$, which will help the regression performance, as we will demonstrate in Section~\ref{sect:results}.

Next, we construct the concatenated spatial filtering matrix $\mathbf{W}^*$ by (\ref{eq:W}), and finally perform the spatial filtering for each EEG trial $\mathbf{X}_n$ by (\ref{eq:Xi}). The complete CSPR-OVR spatial filter for supervised BCI regression problems is summarized in Algorithm~\ref{alg:CSPR}.

\begin{algorithm}[h] 
\KwIn{EEG training examples $(\mathbf{X}_n,y_n)$, where $\mathbf{X}_n\in \mathbb{R}^{C\times S}$, $n=1,...,N$\; \\
\hspace*{10mm} $K$, the number of fuzzy classes for $y_n$\; \\
\hspace*{10mm} $F$, the number of spatial filters for each \\
\hspace*{12mm} fuzzy class.}
\KwOut{Spatially filtered EEG trials $\mathbf{X}_n'\in \mathbb{R}^{KF\times S}$.}
Band-pass filter each $\mathbf{X}_n$ to remove the mean of each channel\;
Compute $\{p_k\}_{k=1,...,K}$ in (\ref{eq:pk})\;
Compute the corresponding percentile values $\{P_k\}_{k=1,...,K}$ for $y_n$\;
Construct the $K$ fuzzy classes as shown in Fig.~\ref{fig:FSs}\;
Compute $\bar{\mathbf{X}}_k$ by (\ref{eq:fP0})\;
Compute $\mathbf{W}_k^*$ by (\ref{eq:CSP})\;
Construct $\mathbf{W}^*$ by (\ref{eq:W})\;
\textbf{Return} $\mathbf{X}_n'$ by (\ref{eq:Xi})
\caption{The CSPR-OVR spatial filter for supervised BCI regression problems.} \label{alg:CSPR}
\end{algorithm}

\subsection{CSPR-OVA}

In (\ref{eq:CSP}) we construct the multi-class CSP using an OVR approach, but it can also be constructed using the following one-versus-all (OVA) approach:
 \begin{align}
    \mathbf{W}_k^*
    =\arg\max\limits_{\mathbf{W}}\frac{\mathrm{Tr}(\mathbf{W}^T\bar{\mathbf{\Sigma}}_k\mathbf{W})}
    {\mathrm{Tr}[\mathbf{W}^T(\sum_{i=1}^K\bar{\mathbf{\Sigma}}_i)\mathbf{W}]} \label{eq:CSP2}
 \end{align}
The only difference between (\ref{eq:CSP2}) and (\ref{eq:CSP}) is that the numerator of (\ref{eq:CSP2}) also includes  the contribution from Class $k$ itself. If we view Class $k$ as the signal of interest, and all other classes as noises, then (\ref{eq:CSP2}) maximizes the signal to signal-plus-noise ratio, as (\ref{eq:xDAWN}) in the xDAWN algorithm.

Equation (\ref{eq:CSP2}) is also a generalized Rayleigh quotient \cite{Golub1996}, and the solution $\mathbf{W}_k^*$ is the concatenation of the $F$ eigenvectors associated with the $F$ largest eigenvalues of the matrix $(\sum_{i=1}^ K\bar{\mathbf{\Sigma}}_i)^{-1}\bar{\mathbf{\Sigma}}_k$. The OVA CSP for classification still uses (\ref{eq:W}) to construct the final spatial filter, and (\ref{eq:Xi}) to perform the filtering.

Using the technique introduced in the previous subsection, we can easily develop the CSPR-OVA spatial filter for BCI regression problems. Its procedure is almost identical to that in Algorithm~\ref{alg:CSPR}. The only difference is that $\mathbf{W}_k^*$ is computed by (\ref{eq:CSP2}) instead of (\ref{eq:CSP}).

\section{Experiments and Data} \label{sect:exp}

This section introduces a PVT experiment that was used to evaluate the performances of the proposed spatial filtering algorithms, the corresponding RS and EEG data preprocessing procedures, and the feature sets.

\subsection{Experiment Setup} \label{sect:PVT}

17 university students (13 males; average age 22.4, standard deviation 1.6) from National Chiao Tung University (NCTU) in Taiwan volunteered to support the data-collection efforts over a 5-month period to study EEG correlates of attention and performance changes under specific conditions of real-world fatigue \cite{Kerick2016}, as determined by the effectiveness score of Readiband \cite{Russell2015}. The voluntary, fully informed consent of the persons used in this research was obtained as required by federal and Army regulations \cite{USArmy,USDoD}. The Institutional Review Board of NCTU approved the experimental protocol.

All participants registered their fatigue levels through a smartphone daily, and received notifications to report for experimental trials when the effectiveness score deemed their conditions fitted the experimental requirement (low fatigue: $>90$; normal: $[70, 90]$; high fatigue: $<70$). Upon completion of the related questionnaires [Karolinska Sleepiness Scale (KSS) \cite{Akerstedt1990}, and electronically-adapted visual analog scale for fatigue (VAS-F) and stress (VAS-S)] and the informed consent form, subjects performed a PVT, a dynamic attention-shifting task, a lane-keeping task, and selected surveys (KSS, VAS-F, VAS-S, state-trait anxiety inventory, and mind-wandering) preceding each condition. EEG data were recorded at 1000 Hz using a 64-channel NeuroScan system. Most participants performed the laboratory experiment thrice in each of the three fatigue states.

In this paper we focus on the PVT \cite{Dinges1985}, which is a sustained-attention task that uses RS to measure the speed with which a subject responds to a visual stimulus. It is widely used, particularly by NASA, for its ease of scoring, simple metrics, convergent validity, and free of learning effects. In our experiment, the PVT was presented on a smartphone with each trial initiated as an empty solid white circle centered on the touchscreen that began to fill in red displayed as a clockwise sweeping motion like the hand of a clock. The sweeping motion was programmed to turn solid red in one second or terminate upon a response by the participants, which required them to tap the touchscreen with the thumb of their dominant hand. The RS was computed as the inverse of the elapsed time between the appearance of the empty solid white circle and the participant's response. Following completion of each trial, the circle went back to solid white until the next trial. Inter-trial intervals consisted of random intervals between 2-10 seconds.

143 sessions of PVT data were collected from the 17 subjects, and each session lasted 10 minutes. Our goal is to predict the RS using a 3-second EEG trial immediately before it.

\subsection{Performance Evaluation Process}

The following procedure was performed to evaluate the performances of different spatial filters:
\begin{enumerate}
\item EEG data preprocessing to suppress artifacts and noises.
\item RS data preprocessing to suppress outliers.
\item 5-fold cross-validation to compute the regression performance for each combination of spatial filters and regression method: first randomly partition the trials into five equal folds; then, use four folds for supervised spatial filtering and regression model training, and the remaining fold for testing; repeat this five times so that every fold is used in testing; finally compute the regression performances in terms of root mean square error (RMSE) and correlation coefficient (CC). Two regression methods were used: LASSO, whose adjustable parameter $\lambda$ was optimized by an inner 5-fold cross-validation on the training dataset, and $k$-nearest neighbors (kNN) regression, where $k=5$.
\item Repeat Step~3 10 times and compute the average regression performance.
\end{enumerate}
More details about the first two steps are given in the next two subsections.

\subsection{EEG Data Preprocessing}

We first downsampled the EEG data to 256 Hz, then epoched them to 3-second trials according to the onset of the PVTs. Let the onset time of the $n$th PVT be $t_n$. Then, the 62-channel EEG trial in $[t_n-3, t_n]$ seconds was used to predict the RS, i.e., $\mathbf{X}_n\in\mathbb{R}^{62\times 768}$. Each trial was then individually filtered by a $[1, 20]$ Hz finite impulse response band-pass filter to make each channel zero-mean and to remove un-useful high frequency components.

Because the inter-trial intervals consisted of random intervals between 2-10 seconds, it's possible that a 3-second EEG trial covers part of data from the previous trial. Additionally, a trial may also contain the EEG oscillations related to motor reaction (tapping the touchscreen) in the previous trial. To remedy these problems, we removed overlapping trials: let the RS of the $n$th trial be $y_n$ (the corresponding response time is $1/y_n$); then, the $n$th trial is removed if $t_n-t_{n-1}<1/y_{n-1}+3$, i.e., when the 3-second EEG data for Trial $n$ overlap with the data and response for the previous trial.

\subsection{RS Data Preprocessing}

\begin{figure}[htpb]\centering
\includegraphics[width=\linewidth,clip]{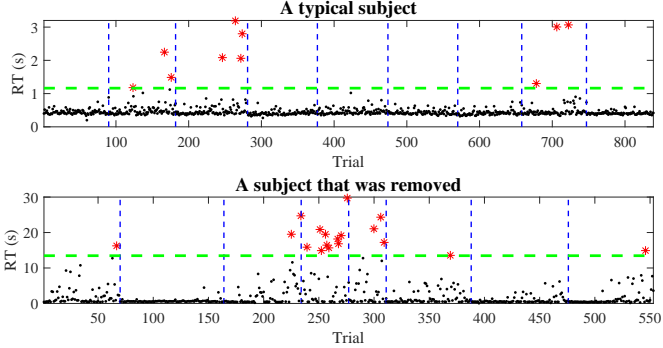}
\caption{Response times for a typical subject (top panel) and a subject with possible data recording issues (bottom panel). The green line is the threshold, and the red stars are response times above the threshold, which will be brought to the threshold.} \label{fig:RTs}
\end{figure}

The raw response times for two subjects are shown in Fig.~\ref{fig:RTs}. The top panel is from a typical subject, whose response times were mostly shorter than 1 second. The lower panel is from a subject with possible data recording issues, because lots of response times were longer than 5 seconds, which are highly unlikely in practice. So we excluded that subject from consideration in this paper, and only used the remaining 16 subjects.

\begin{figure}[htpb]\centering
\includegraphics[width=\linewidth,clip]{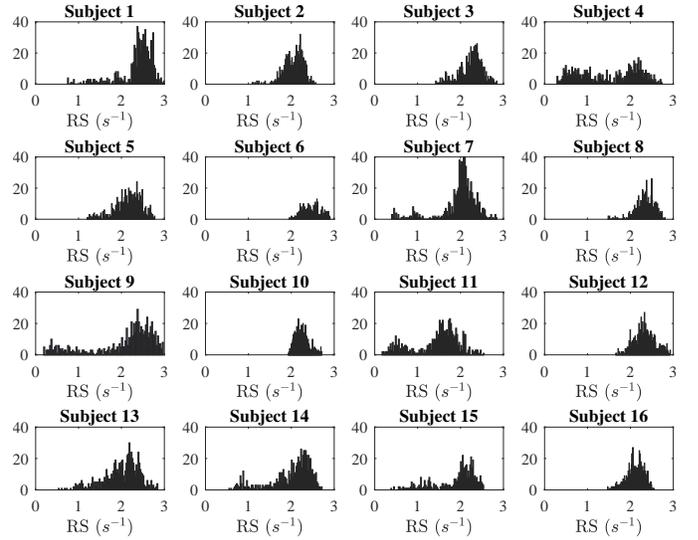} \caption{Distributions of the preprocessed RSs for the 16 subjects.} \label{fig:RSs}
\end{figure}

As shown in Fig.~\ref{fig:RTs}, the response times were very noisy, and there were obvious outliers. It is very important to suppress the outliers and noises so that the performances of different algorithms can be more accurately compared. In addition to the step in the previous subsection to remove overlapping trials, we also employed the following 2-step procedure for response time preprocessing:
\begin{enumerate}
\item \emph{Outlier thresholding}, which aimed to suppress abnormally large response times. First, a threshold $\theta=m_y+3\sigma_y$ was computed for each subject, where $m_y$ is the mean response time from all sessions of that subject, and $\sigma_y$ is the corresponding standard deviation. Then, all response times larger than $\theta$ were replaced by $\theta$. Note that the threshold was different for different subjects.
\item \emph{Moving average smoothing}, which replaced each response time by the average response time during a 60 seconds moving window centered at the onset of the corresponding PVT to suppress noises.
\end{enumerate}
We then computed the RS as the inverse of the RT. The RSs for the 16 subjects are shown in Fig.~\ref{fig:RSs}. Observe that they are roughly in the same range, and many of them are approximately Gaussian.

\subsection{Feature Extraction}

We extracted the following four feature sets for each preprocessed EEG trial:
\begin{itemize}
\item \texttt{Raw}: \emph{Theta and Alpha powerband features from the band-pass filtered EEG trials.} We computed the average power spectral density (PSD) in the Theta band (4-8 Hz) and Alpha band (8-13 Hz) for each channel using Welch's method \cite{Welch1967}, and converted these $62\times 2=124$ band powers to dBs as our features.
\item \texttt{CAR}: \emph{Theta and Alpha powerband features from EEG trials filtered by CAR.} This procedure was almost identical to \texttt{Raw}, except that the band-pass filtered EEG trials were also spatially filtered by CAR before the $62\times 2=124$ powerband features were computed. CAR is one of the most commonly used spatial filters for EEG, and \cite{McFarland1997} showed that it helped improve EEG classification performance. It simply removes the mean of all channels from each channel.
\item \texttt{OVR}: \emph{Theta and Alpha powerband features from EEG trials filtered by CSPR-OVR.} This procedure was almost identical to \texttt{CAR}, except that the CAR filter was replaced by CSPR-OVR. We used 3 fuzzy classes for the RSs, and 21 spatial filters\footnote{We used 21 spatial filters here so that the filtered signals had roughly the same dimensionality as the original signals, which ensured fair performance comparison. In Section~\ref{sect:sensitivity} we also performed sensitivity analysis on the number of spatial filters.} for each fuzzy class, so that the spatially filtered signals had dimensionality $63\times 1280$, roughly the same as the dimensionality of the original signals. We then extracted $63\times 2=126$ band power features for each trial.
\item \texttt{OVA}: \emph{Theta and Alpha powerband features from EEG trials filtered by CSPR-OVA.} This procedure was also almost identical to \texttt{CAR}, except that the spatial filtering was performed by CSPR-OVA instead of CAR. There were also $63\times 2=126$ band power features for each trial.
\end{itemize}

\section{Experimental Results} \label{sect:results}

This section compares the informativeness of the features in \texttt{Raw}, \texttt{CAR}, \texttt{OVR} and \texttt{OVA}, presents the regression performances, and also performs parameter sensitivity analysis for Algorithm~1.

\subsection{Informativeness of the Features}

Before studying the regression performances, it is important to check if the extracted features in \texttt{Raw}, \texttt{CAR}, \texttt{OVR} and \texttt{OVA} are indeed meaningful. We picked a typical subject, partitioned his data random into 50\% training and 50\% testing, and extracted \texttt{Raw} and \texttt{CAR}. We then designed the spatial filters using CSPR-OVR and CSPR-OVA on the training data, and extracted the corresponding \texttt{OVR} and \texttt{OVA}. For each feature set, we identified the top three channels that had the maximum correlation with the RS using the training data, and also computed the corresponding correlation coefficients for the testing data.

\begin{figure}[htpb]\centering
\includegraphics[width=\linewidth,clip]{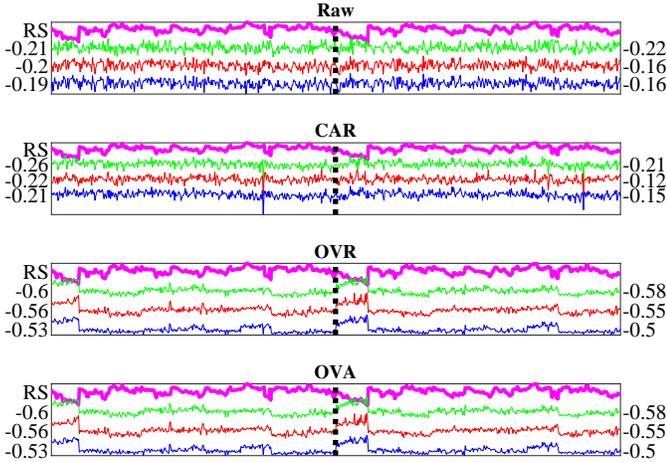}
\caption{Powerband features from different feature extraction methods, and the corresponding training and testing CCs with the RS.} \label{fig:corr}
\end{figure}

The results are shown in Fig.~\ref{fig:corr}, where in each subfigure the data on the left of the black dotted line were used for training, and the right for testing. The top thick curve is the RS, and the bottom three curves are the maximally correlated features (note that good features are negatively correlated with the RS) identified from the training data. The training and testing correlation coefficients are shown on the left and right of the corresponding channel, respectively. Observe that the features from \texttt{CAR} had slightly better correlations with the RS in training than those from \texttt{Raw}, but not necessarily in testing. However, the features from \texttt{OVR} and \texttt{OVA} had much higher training and testing correlations to the RS than those from \texttt{Raw} and \texttt{CAR}, suggesting that CSPR-OVR and CSPR-OVA can indeed increase the signal quality. The reason is: if we view Class $k$ as the signal of interest, and all other classes as noises, then CSPR-OVR in (\ref{eq:CSP}) enhances the signal to noise ratio of the EEG signal, and CSPR-OVA in (\ref{eq:CSP2}) enhances the signal to signal-plus-noise ratio.

\subsection{Regression Performance Comparison}

\begin{figure*}[htpb]\centering
\includegraphics[width=.8\linewidth,clip]{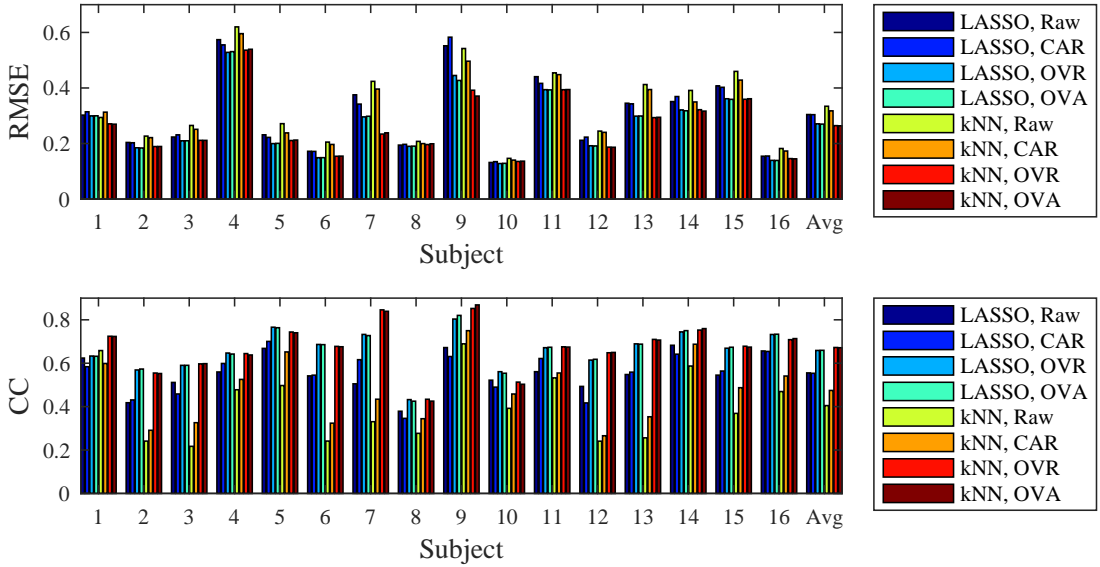}
\caption{RMSEs and CCs of the eight approaches on the 16 subjects. } \label{fig:perf}
\end{figure*}

The RMSEs and CCs of LASSO and kNN using the four feature sets are shown in Fig.~\ref{fig:perf} for the 16 subjects. Recall that for each subject the feature extraction methods were run 10 times, each with randomly partitioned training and testing data, and the average regression performances are shown here. The average RMSEs and CCs across all subjects are also shown in the last group of each panel. Observe that \texttt{CAR} had comparable or slightly better performance than \texttt{Raw}. Regardless of which regression algorithm was used, generally \texttt{OVR} and \texttt{OVA} had similar performance, and both of them achieved much smaller RMSEs and much larger CCs than \texttt{Raw} and \texttt{CAR}, suggesting that our extension of CSP from supervised classification to supervised regression can indeed improve the regression performance. Finally, LASSO had better performance than kNN on \texttt{Raw} and \texttt{CAR}, but kNN became better on \texttt{OVR} and \texttt{OVA}.

\begin{figure*}[htpb]\centering
\includegraphics[width=.8\linewidth,clip]{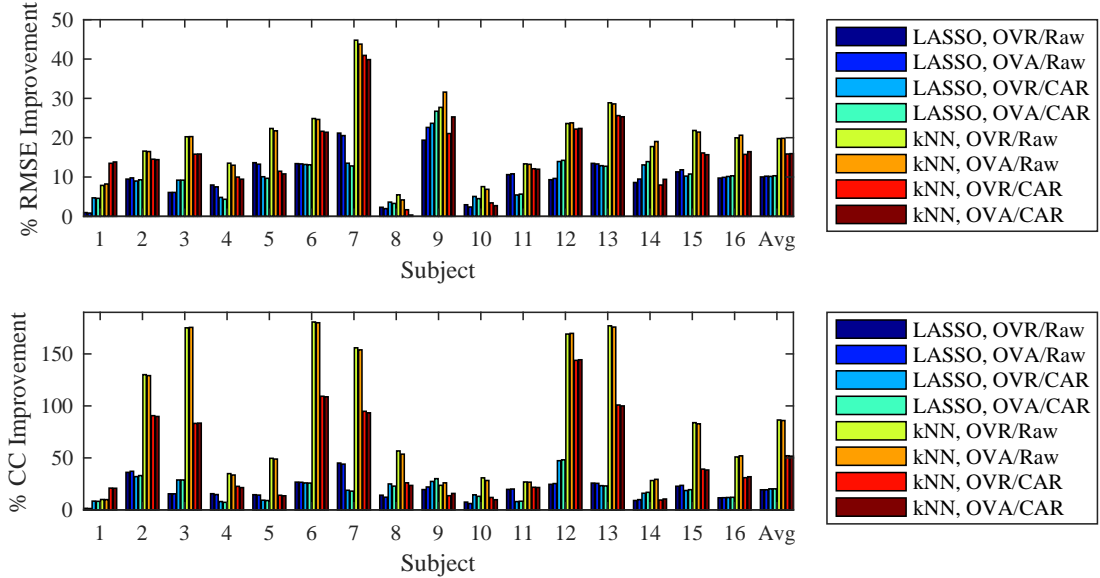}
\caption{Pairwise percentage performance improvement of the algorithms on the 16 subjects. } \label{fig:prc}
\end{figure*}

The corresponding percentage performance improvements of LASSO and kNN using the four feature sets are shown in Fig.~\ref{fig:prc}, where the legend ``\texttt{LASSO,OVR}/\texttt{Raw}" means the percentage performance improvement of LASSO on \texttt{OVR} over LASSO on \texttt{Raw}, and other legends should be interpreted in a similar manner. For both LASSO and kNN, \texttt{OVR} and \texttt{OVA} achieved similar performance improvements over \texttt{Raw}, and also over \texttt{CAR}. For LASSO, on average \texttt{OVR} had $10.02\%$ smaller RMSE than \texttt{Raw}, and $19.39\%$ larger CC. For kNN, on average \texttt{OVR} had $19.77\%$ smaller RMSE than \texttt{Raw}, and $86.47\%$ larger CC.

We also performed a two-way Analysis of Variance (ANOVA) for different regression algorithms to check if the RMSE and CC differences among the four feature sets were statistically significant, by setting the subjects as a random effect. The results are shown in Table~\ref{tab:ANOVA}, which indicated that there were statistically significant differences in both RMSEs and CCs among different feature sets for both LASSO and kNN.

\begin{table}[!ht] \centering \setlength{\tabcolsep}{2mm}
\caption{$p$-values of two-way ANOVA tests for $\{\texttt{Raw},\ \texttt{CAR},\ \texttt{OVR},\ \texttt{OVA}\}$.}   \label{tab:ANOVA}
\begin{tabular}{l|cc||cc}   \hline
&\multicolumn{2}{c||}{LASSO} & \multicolumn{2}{c}{kNN} \\ \hline
&RMSE & CC & RMSE & CC \\ \hline
   $p$  & $\mathbf{.0061}$ & $\mathbf{.0000}$ & $\mathbf{.0000}$ &  $\mathbf{.0000}$\\ \hline
\end{tabular}
\end{table}

Then, non-parametric multiple comparison tests based on Dunn's procedure \cite{Dunn1961,Dunn1964} were used to determine if the difference between any pair of algorithms was statistically significant, with a $p$-value correction using the False Discovery Rate method \cite{Benjamini1995}. The $p$-values are shown in Table~\ref{tab:Dunn1}, where the statistically significant ones are marked in bold. Table~\ref{tab:Dunn1} shows that, except for the CC of kNN, generally there was no statistically significant difference between \texttt{Raw} and \texttt{CAR}. However, for both LASSO and kNN, the RMSE and CC differences between $\{\texttt{OVR},\ \texttt{OVA}\}$ and $\{\texttt{Raw},\ \texttt{CAR}\}$ were always statistically significant. In all cases, there were no statistically significant differences between \texttt{OVR} and \texttt{OVA}.

\begin{table}[!ht] \centering \setlength{\tabcolsep}{.2mm}
\caption{$p$-values of non-parametric multiple comparison for $\{\texttt{Raw},\ \texttt{CAR}, \ \texttt{OVR}, \ \texttt{OVA}\}$.}   \label{tab:Dunn1}
\begin{tabular}{l|ccc|ccc|ccc|ccc}   \hline
&\multicolumn{6}{c|}{LASSO} & \multicolumn{6}{c}{kNN} \\ \hline
&\multicolumn{3}{c|}{RMSE} &\multicolumn{3}{c|}{CC} & \multicolumn{3}{c|}{RMSE} & \multicolumn{3}{c}{CC} \\ \hline
     &  \texttt{Raw} & \texttt{CAR} & \texttt{OVR} &  \texttt{Raw} & \texttt{CAR} & \texttt{OVR}  &  \texttt{Raw} & \texttt{CAR} & \texttt{OVR} &  \texttt{Raw} & \texttt{CAR} & \texttt{OVR}  \\ \hline
\texttt{CAR} &  .5883       &              &     &.3374         &              &      &.1437         &       &               & \textbf{.0009} & &  \\
\texttt{OVR} &\textbf{.0063}&\textbf{.0034}&     &\textbf{.0000}&\textbf{.0000}&      &\textbf{.0000}& \textbf{.0001}&  & \textbf{.0000}& \textbf{.0000} &\\
\texttt{OVA} &\textbf{.0122}&\textbf{.0044}&.4960&\textbf{.0000}&\textbf{.0000}&.4970 &\textbf{.0000}& \textbf{.0001}&.4937& \textbf{.0000}& \textbf{.0000} &.4741\\ \hline
\end{tabular}
\end{table}

\subsection{Parameter Sensitivity Analysis} \label{sect:sensitivity}

There are two adjustable parameters in CSPR-OVR: $K$, the number of fuzzy classes for the RSs, and $F$, the number of spatial filters for each fuzzy class. In this subsection we study the sensitivity of the regression performance to these  two parameters.

\begin{figure}[htpb]\centering
\subfigure[]{\label{fig:RMSEnFSs}     \includegraphics[width=\linewidth,clip]{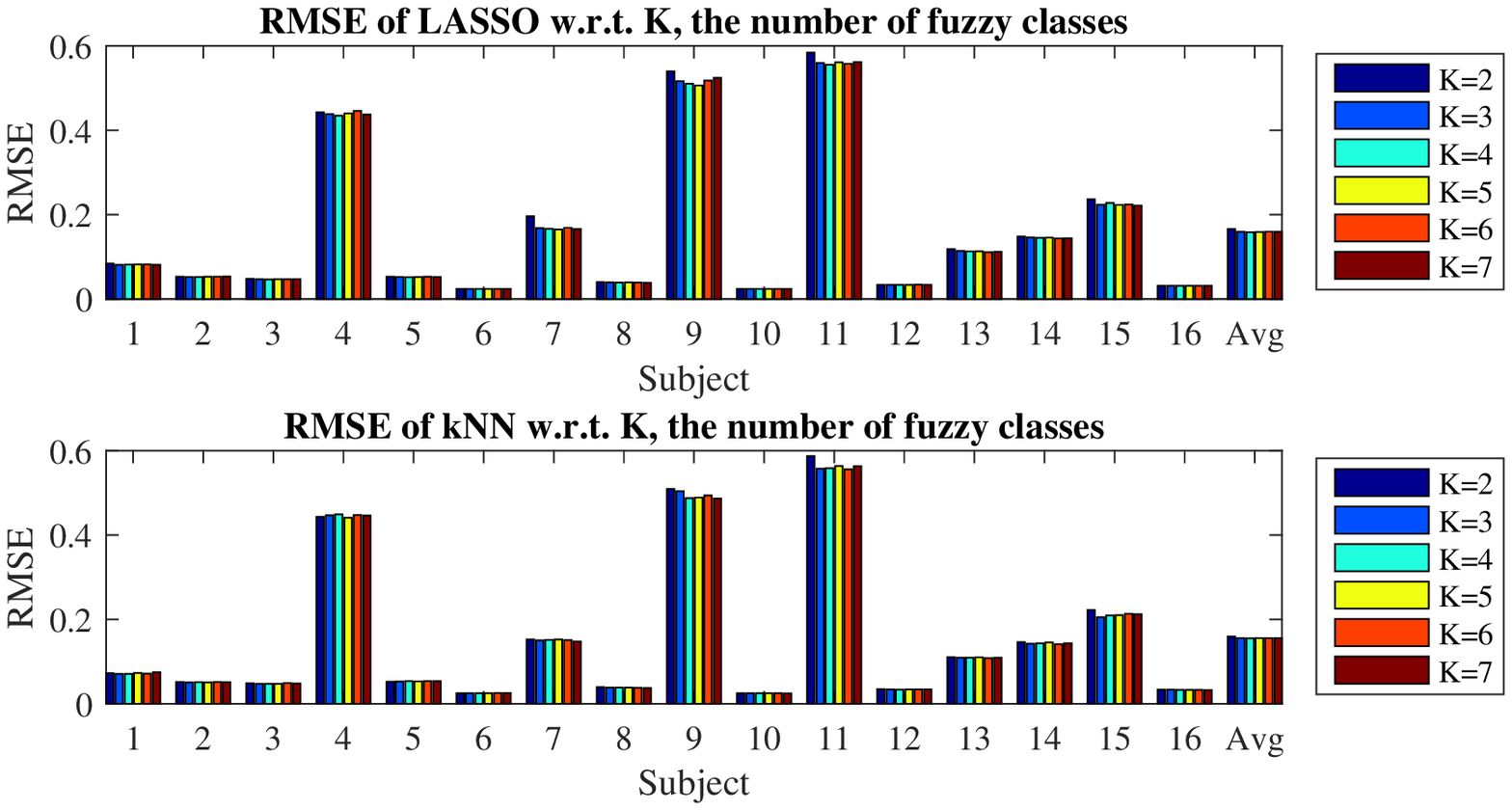}}
\subfigure[]{\label{fig:CCnFSs}     \includegraphics[width=\linewidth,clip]{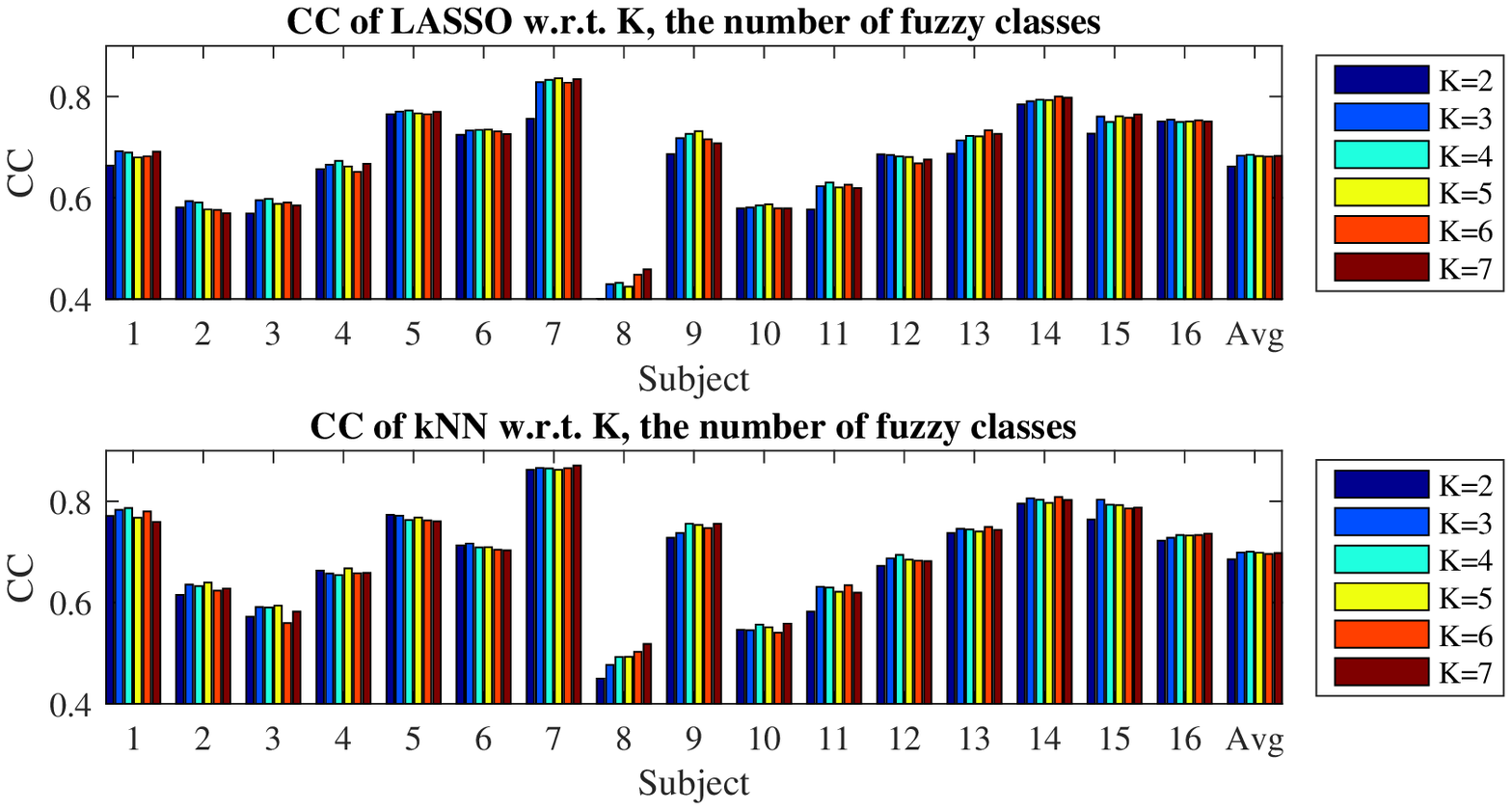}}
\caption{(a) RMSEs and (b) CCs of LASSO and kNN with respect to $K$, the number of fuzzy classes in Algorithm~1.} \label{fig:nFSs}
\end{figure}

The regression performances for $K=\{2, 3, 4, 5, 6, 7\}$ ($F$ was fixed to be 21) are shown in Fig.~\ref{fig:nFSs}. Algorithm~1 was repeated five times, each with a random partition of training and testing data, and the average regression results are shown. For both LASSO and kNN, on average $K=2$ gave worst performance, but $K=\{3,4,5,6,7\}$ resulted in roughly the same RMSE and CC. Hence, $K=3$ seems to be a good compromise between performance and computational cost.

\begin{figure}[htpb]\centering
\subfigure[]{\label{fig:RMSEnFSs}     \includegraphics[width=\linewidth,clip]{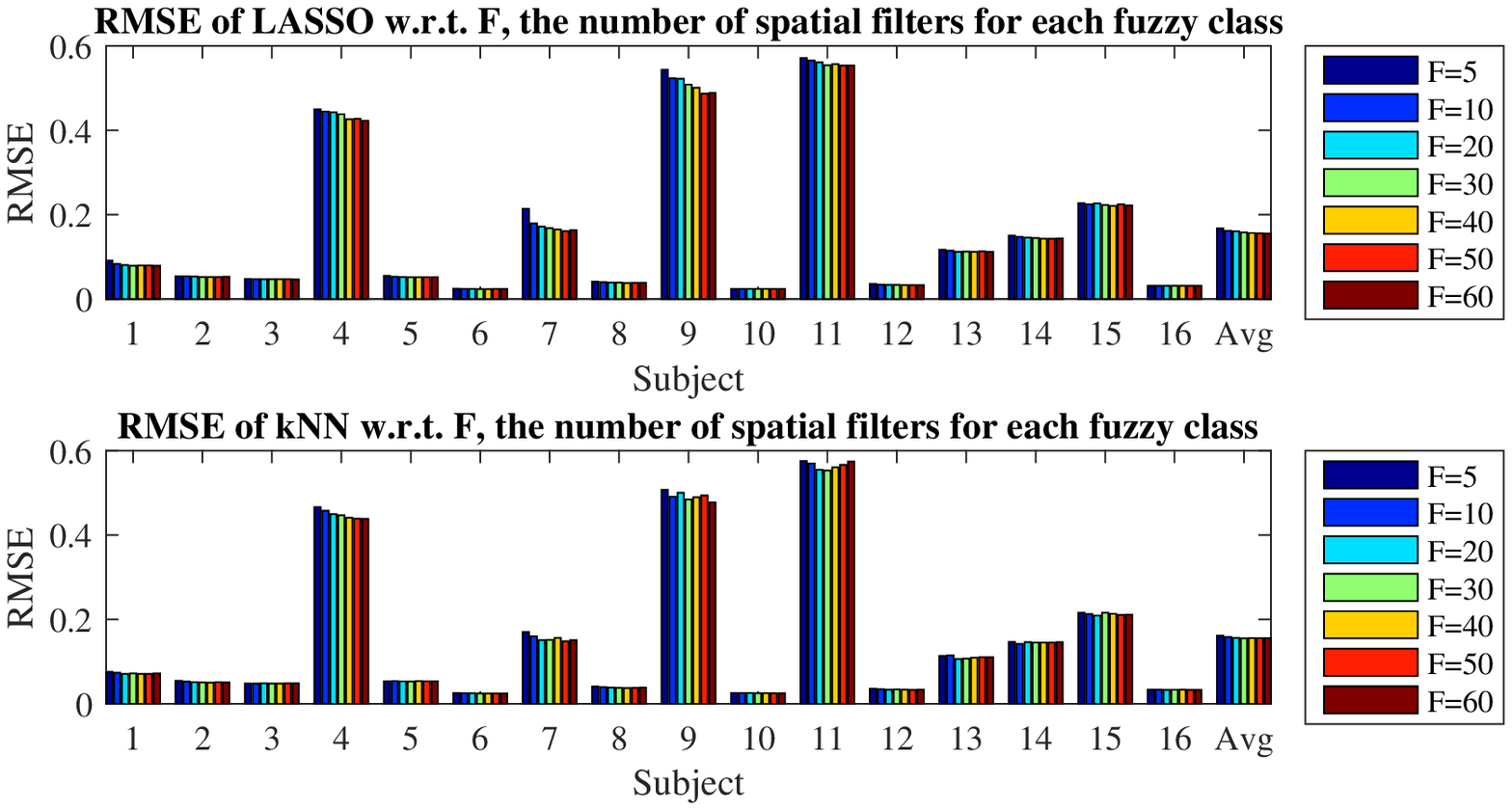}}
\subfigure[]{\label{fig:CCnFSs}     \includegraphics[width=\linewidth,clip]{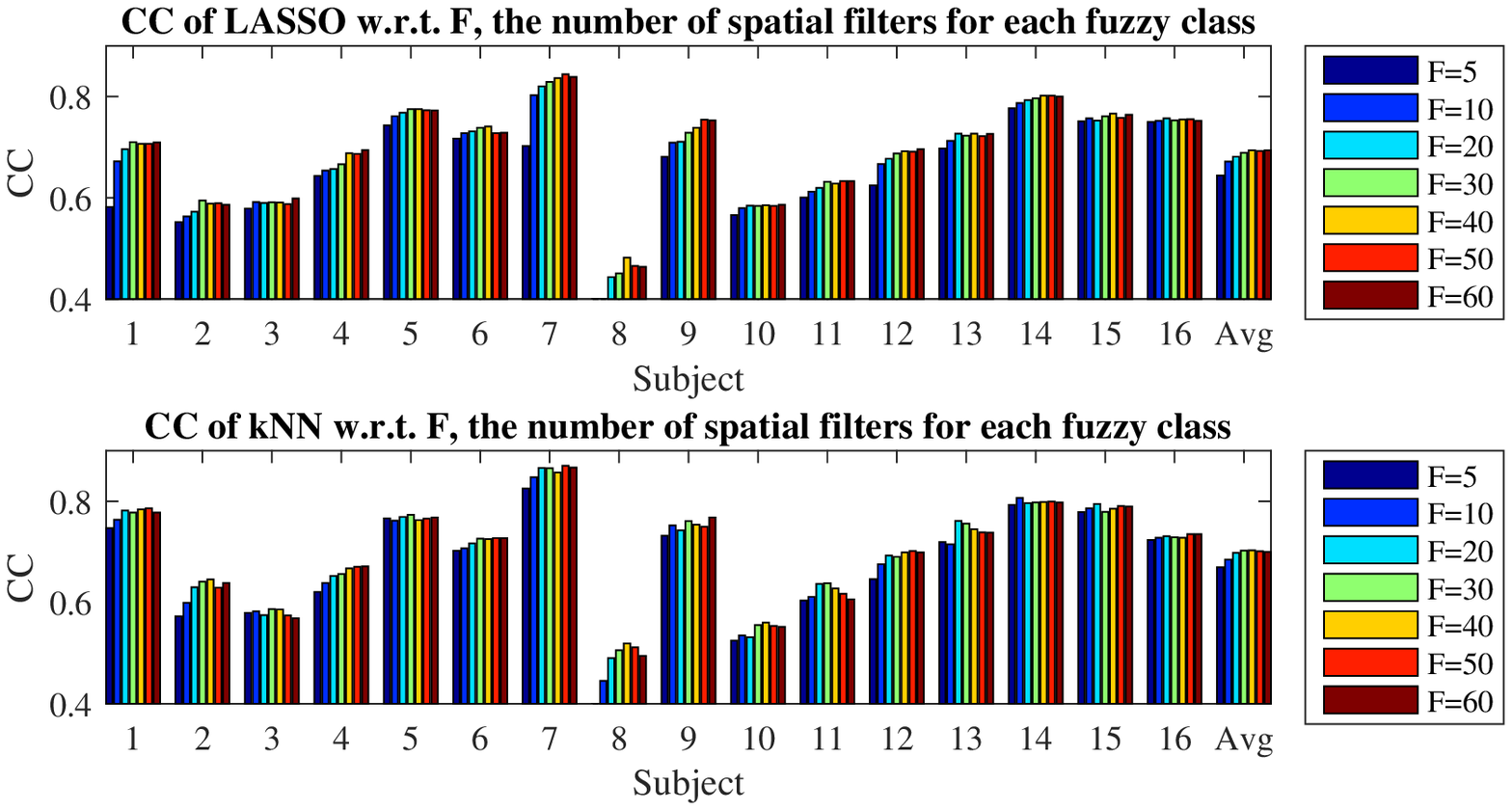}}
\caption{(a) RMSEs and (b) CCs of LASSO and kNN with respect to $F$, the number of spatial filters for each fuzzy class in Algorithm~1.} \label{fig:nFilters}
\end{figure}

The regression performances for $F=\{5, 10, 20, 30, 40, 50, 60\}$ ($K$ was fixed to be 3) are shown in Fig.~\ref{fig:nFilters}. Algorithm~1 was again repeated five times, and the average regression results are shown. For both LASSO and kNN, generally a larger $F$ resulted in a smaller RMSE and a larger CC, but the performance may reach a plateau at a certain $F$. Also, a larger $F$ means heavier computational cost, which should be taken into consideration in choosing $F$. For the PVT experiment, $F\in[20,30]$ seemed to achieve a good compromise between performance and computational cost.

\subsection{Different Fuzzy Set Shapes}

\begin{figure}[htpb]\centering
\includegraphics[width=8cm,clip]{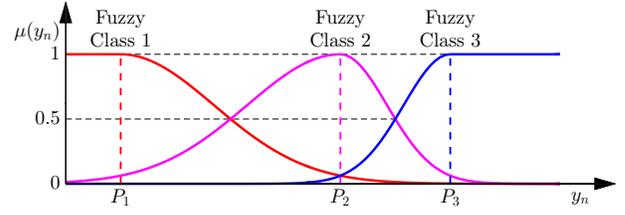} \caption{The three fuzzy classes for $y_n$, when Gaussian fuzzy sets are used.} \label{fig:GaussianMFs}
\end{figure}

In Section~\ref{sect:Filter} we used triangular fuzzy sets for simplicity, but other shapes can also be used. Fig.~\ref{fig:GaussianMFs} illustrates how Gaussian fuzzy sets can be designed here: the center of the $k$th Gaussian fuzzy class is at $P_k$ [computed from (\ref{eq:pk})], and the spread is specially designed so that two adjacent fuzzy sets intersect at the midpoint with membership grade 0.5. As a result, generally the Gaussian fuzzy classes are not symmetric.

\begin{figure}[htpb]\centering
\includegraphics[width=\linewidth,clip]{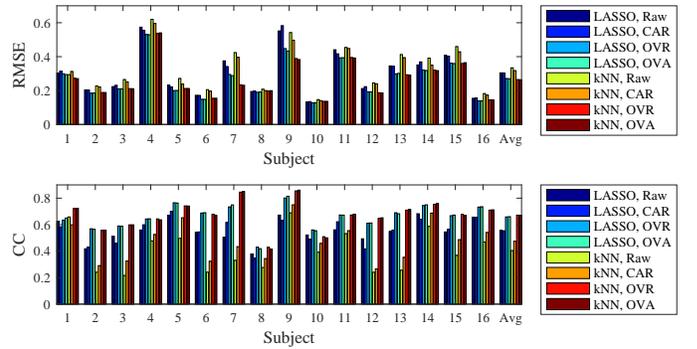}
\caption{RMSEs and CCs of the eight approaches on the 16 subjects, when the three Gaussian fuzzy sets in Fig.~\ref{fig:GaussianMFs} are used in CSPR-OVR and CSPR-OVA. } \label{fig:perfGaussian}
\end{figure}

When the Gaussian fuzzy classes in Fig.~\ref{fig:GaussianMFs} are used in CSPR-OVR and CSPR-OVA, the results are shown in Fig.~\ref{fig:perfGaussian}, which are almost identical to those obtained from triangular fuzzy sets (Fig.~\ref{fig:perf}).

\subsection{Robustness to Noise}

It is also important to study the robustness of different spatial filters to noises. According to \cite{Zhu2004}, there are two types of noises: \emph{class noise}, which is the noise on the model outputs, and \emph{attribute noise}, which is the noise on the model inputs. In this subsection we focus on the attribute noise.

\begin{figure}[htpb]\centering
\includegraphics[clip,width=\linewidth]{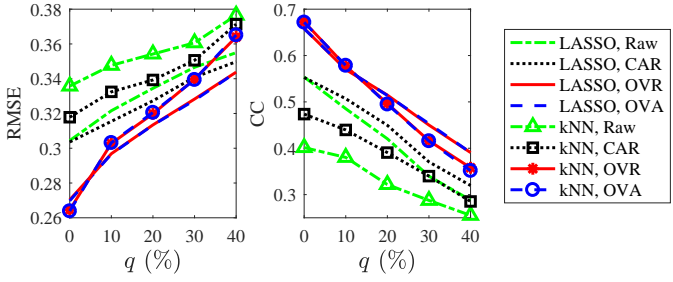} \caption{Average RMSEs and CCs of the eight approaches wrt different attribute noise levels.} \label{fig:noises}
\end{figure}

As in \cite{Zhu2004}, for each model input, we randomly replaced $q\%$ ($q=0,\,10,...,40$) of all trials from a subject with a uniform noise between its minimum and maximum values. After this was done for both the training and testing data, we extracted feature sets \texttt{Raw}, \texttt{CAR}, \texttt{OVR} and \texttt{OVA}, and trained LASSO and kNN, on the corrupted training data. We then tested their performances on the corrupted testing data. The results are shown in Fig.~\ref{fig:noises}. Generally, as the noise level increased, the performances decreased, which is intuitive. However, \texttt{OVR} and \texttt{OVA} achieved better RMSEs and CCs than \texttt{Raw} and \texttt{CAR} at almost all noise levels, suggesting that it is still beneficial to use CSPR-OVR and CSPR-OVA even under high attribute noise.

\subsection{Computational cost}

\begin{figure}[htpb]\centering
\includegraphics[clip,width=.8\linewidth]{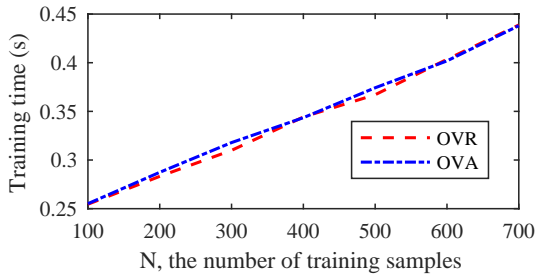} \caption{The training time of CSPR-OVR and CSPR-OVA wrt $N$.} \label{fig:compCost}
\end{figure}

Observe from Algorithm~1 that in training CSPR-OVR needs to perform a matrix inversion and an eigen-decomposition to compute $\mathbf{W}^*$; however, once the training is done, the filtering of new EEG trials can be conducted very efficiently by a simple matrix multiplication [see (\ref{eq:Xi})]. Let $N$ be the number of training samples. Then, the actual training time of CSPR-OVR and CSPR-OVA increased linearly with $N$, as shown in Fig.~\ref{fig:compCost}. The platform was a Dell XPS15 laptop (Intel i7-6700HQ CPU @2.60GHz, 16 GB memory) running Windows 10 Pro 64-bit and Matlab 2016b. A least squares curve fit shows that the training time is $0.2216+0.0003N$ seconds, which should not be a problem for a practical $N$.

\section{Discussions and Future Research} \label{sect:discussions}

Recall that 5-fold cross-validation was used in the performance evaluation in the previous section, i.e., we concatenated the nine-session data from the same subject, randomly partitioned them into five equal-length folds, and then used four folds for training and the remaining one for testing. So, the training and testing folds contained data from the same sessions. This is equivalent to the case that we label some session-specific data in offline regression. Our results showed that in this case CSPR-OVR and CSPR-OVA can significantly improve the regression performance.

To avoid the use of session-specific data, we also investigated a different validation method: leave-one-session-out validation, in which for each subject we trained the spatial filters using eight sessions and tested them on the remaining session. Interestingly, all four feature sets and both regression models achieved very poor performance here. The reasons are: 1) we need a proper way to normalize the RSs from different sessions, as done for the response times in \cite{Hu2015}; and, 2) there is large intra-subject variation, meaning that the EEG responses for the same subject vary at different times (recall that these nine sessions were collected at different days); so, the patterns learned from previous sessions become obsolete for the new session, and hence spatial filtering alone does not help. However, our previous research \cite{drwuTHMS2016,drwuTFS2016,drwuSMLR2016} has shown that transfer learning can cope well with the inter-subject variation (individual differences) in both classification and regression problems, and we conjecture that it can also handle the intra-subject variation. One of our future research directions is to demonstrate the performance of CSPR-OVR and CSPR-OVA in a transfer learning framework to individualize a generalized model for regression problems, as done in \cite{Johnson2011,Stikic2011} for EEG-based cognitive performance classification.

Another direction of our future research will apply CSPR-OVR and CSPR-OVA to other important EEG-based regression problems, e.g., drowsiness (or alertness) estimation during driving, and integrate it with more sophisticated feature extraction approaches, e.g., Riemannian geometry \cite{Congedo2013}, for better regression performance.

\section{Conclusions} \label{sect:conclusions}

EEG signals are easily contaminated by artifacts and noises, so preprocessing is needed before they are fed into a machine learning algorithm in BCI. Spatial filters, e.g., ICA, xDAWN, CSP and CCA, have been widely used to increase the EEG signal quality for classification problems, but their applications in BCI regression problems have been very limited. In this paper, we have proposed two CSP filters for EEG-based regression problems in BCI, which were extended from the CSP filter for classification, by making use of fuzzy sets. Extensive experimental results on EEG-based RS estimation from a large-scale study, which collected 143 sessions of PVT data from 17 subjects during a 5-month period, demonstrated that our proposed spatial filters can significantly increase the EEG signal quality. When used in LASSO and kNN, the spatial filters can reduce the estimation RMSE by $10.02-19.77\%$, and at the same time increase the CC by $19.39-86.47\%$.

\section*{Acknowledgement}

Research was sponsored by the U.S. Army Research Laboratory and was accomplished under Cooperative Agreement Numbers W911NF-10-2-0022 and W911NF-10-D-0002/TO 0023. The views and the conclusions contained in this document are those of the authors and should not be interpreted as representing the official policies, either expressed or implied, of the U.S. Army Research Laboratory or the U.S. Government. This work was also partially supported by the Australian Research Council (ARC) under discovery grant DP150101645.



\begin{thebibliography}{10}
\providecommand{\url}[1]{#1}
\csname url@samestyle\endcsname
\providecommand{\newblock}{\relax}
\providecommand{\bibinfo}[2]{#2}
\providecommand{\BIBentrySTDinterwordspacing}{\spaceskip=0pt\relax}
\providecommand{\BIBentryALTinterwordstretchfactor}{4}
\providecommand{\BIBentryALTinterwordspacing}{\spaceskip=\fontdimen2\font plus
\BIBentryALTinterwordstretchfactor\fontdimen3\font minus
  \fontdimen4\font\relax}
\providecommand{\BIBforeignlanguage}[2]{{%
\expandafter\ifx\csname l@#1\endcsname\relax
\typeout{** WARNING: IEEEtranS.bst: No hyphenation pattern has been}%
\typeout{** loaded for the language `#1'. Using the pattern for}%
\typeout{** the default language instead.}%
\else
\language=\csname l@#1\endcsname
\fi
#2}}
\providecommand{\BIBdecl}{\relax}
\BIBdecl

\bibitem{Akerstedt1990}
T.~Akerstedt and M.~Gillberg, ``Subjective and objective sleepiness in the
  active individual,'' \emph{International Journal of Neuroscience}, vol.~52,
  no. 1-2, pp. 29--37, 1990.

\bibitem{Barachant2014b}
\BIBentryALTinterwordspacing
A.~Barachant. (2014) {MEG} decoding using {R}iemannian geometry and
  unsupervised classification. Accessed: 8/17/2016. [Online]. Available:
  \url{http://alexandre.barachant.org/wp-content/uploads/2014/08/documentation.pdf.}
\BIBentrySTDinterwordspacing

\bibitem{Benjamini1995}
Y.~Benjamini and Y.~Hochberg, ``Controlling the false discovery rate: A
  practical and powerful approach to multiple testing,'' \emph{Journal of the
  Royal Statistical Society, Series B (Methodological)}, vol.~57, pp. 289--300,
  1995.

\bibitem{Bigdely-Shamlo2015}
N.~Bigdely-Shamlo, T.~Mullen, C.~Kothe, K.-M. Su, and K.~A. Robbins, ``The
  {PREP} pipeline: standardized preprocessing for large-scale {EEG} analysis,''
  \emph{Frontiers in Neuroinformatics}, vol.~9, 2015.

\bibitem{Bin2011}
G.~Bin, X.~Gao, Y.~Wang, Y.~Li, B.~Hong, and S.~Gao, ``A high-speed {BCI} based
  on code modulation {VEP},'' \emph{Journal of neural engineering}, vol.~8,
  no.~2, 2011.

\bibitem{Bin2009a}
G.~Bin, X.~Gao, Z.~Yan, B.~Hong, and S.~Gao, ``An online multi-channel
  {SSVEP}-based brain-computer interface using a canonical correlation analysis
  method,'' \emph{Journal of neural engineering}, vol.~6, no.~4, 2009.

\bibitem{Blankertz2008}
B.~Blankertz, R.~Tomioka, S.~Lemm, M.~Kawanabe, and K.~R. Muller, ``Optimizing
  spatial filters for robust {EEG} single-trial analysis,'' \emph{{IEEE} Signal
  Processing Magazine}, vol.~25, no.~1, pp. 41--56, 2008.

\bibitem{Congedo2013}
M.~Congedo, A.~Barachant, and A.~Andreev, ``A new generation of brain-computer
  interface based on {R}iemannian geometry,'' \emph{arXiv: 1310.8115}, 2013.

\bibitem{Delorme2004}
A.~Delorme and S.~Makeig, ``{EEGLAB}: an open source toolbox for analysis of
  single-trial {EEG} dynamics including independent component analysis,''
  \emph{Journal of Neuroscience Methods}, vol. 134, pp. 9--21, 2004.

\bibitem{Dinges1985}
D.~F. Dinges and J.~W. Powell, ``Microcomputer analyses of performance on a
  portable, simple visual {RT} task during sustained operations,''
  \emph{Behavior research methods, instruments, \& computers}, vol.~17, no.~6,
  pp. 652--655, 1985.

\bibitem{Dornhege2004}
G.~Dornhege, G.~C. B.~Blankertz, and K.-R. Muller, ``Boosting bit rates in
  non-invasive {EEG} single-trial classifications by feature combination and
  multi-class paradigms,'' \emph{{IEEE} Trans. on Biomedical Engineering},
  vol.~51, no.~6, pp. 993--1002, 2004.

\bibitem{Dunn1961}
O.~Dunn, ``Multiple comparisons among means,'' \emph{Journal of the American
  Statistical Association}, vol.~56, pp. 62--64, 1961.

\bibitem{Dunn1964}
------, ``Multiple comparisons using rank sums,'' \emph{Technometrics}, vol.~6,
  pp. 214--252, 1964.

\bibitem{Golub1996}
G.~H. Golub and C.~F.~V. Loan, \emph{Matrix Computation}, 3rd~ed.\hskip 1em
  plus 0.5em minus 0.4em\relax Baltimore, MD: The Johns Hopkins University
  Press, 1996.

\bibitem{Hotelling1936}
H.~Hotelling, ``Relations between two sets of variates,'' \emph{Biometrika},
  vol.~28, no. 3/4, pp. 321--377, 1936.

\bibitem{Hu2015}
Z.~Hu, Y.~Sun, J.~Lim, N.~Thakor, and A.~Bezerianos, ``Investigating the
  correlation between the neural activity and task performance in a psychomotor
  vigilance test,'' in \emph{Proc. 37th Annual Int'l Conf. of the {IEEE}
  Engineering in Medicine and Biology Society ({EMBC})}, Milan, Italy, August
  2015, pp. 4725--4728.

\bibitem{Hyvarinen2000}
A.~Hyvarinen and E.~Oja, ``Independent component analysis: algorithms and
  applications,'' \emph{Neural networks}, vol.~13, no.~4, pp. 411--430, 2000.

\bibitem{Johnson2011}
R.~R. Johnson, D.~P. Popovic, R.~E.~O. andMaja Stikic, D.~J. Levendowski, and
  C.~Berka, ``Drowsiness/alertness algorithm development and validation using
  synchronized {EEG} and cognitive performance to individualize a generalized
  model,'' \emph{Biological Psychology}, vol.~87, p. 241–250, 2011.

\bibitem{Jolliffe2002}
I.~Jolliffe, \emph{Principal component analysis}.\hskip 1em plus 0.5em minus
  0.4em\relax Wiley Online Library, 2002.

\bibitem{Jung2000}
T.-P. Jung, S.~Makeig, C.~Humphries, T.-W. Lee, M.~J. Mckeown, V.~Iragui, and
  T.~J. Sejnowski, ``Removing electroencephalographic artifacts by blind source
  separation,'' \emph{Psychophysiology}, vol.~37, no.~2, pp. 163--178, 2000.

\bibitem{Kerick2016}
S.~Kerick, C.-H. Chuang, J.-T. King, T.-P. Jung, J.~Brooks, B.~T. Files,
  K.~McDowell, and C.-T. Lin, ``Inter- and intra-individual variations in
  sleep, subjective fatigue, and vigilance task performance of students in
  their real-world environments over extended periods,'' 2016, submitted.

\bibitem{Klir1995}
G.~J. Klir and B.~Yuan, \emph{Fuzzy Sets and Fuzzy Logic: {T}heory and
  Applications}.\hskip 1em plus 0.5em minus 0.4em\relax Upper Saddle River, NJ:
  Prentice-Hall, 1995.

\bibitem{Lagerlund1997}
T.~D. Lagerlund, F.~W. Sharbrough, and N.~E. Busacker, ``Spatial filtering of
  multichannel electroencephalographic recordings through principal component
  analysis by singular value decomposition,'' \emph{Journal of Clinical
  Neurophysiology}, vol.~14, no.~1, pp. 73--82, 1997.

\bibitem{Lance2012}
B.~J. Lance, S.~E. Kerick, A.~J. Ries, K.~S. Oie, and K.~McDowell,
  ``Brain-computer interface technologies in the coming decades,'' \emph{Proc.
  of the {IEEE}}, vol. 100, no.~3, pp. 1585--1599, 2012.

\bibitem{Liao2012}
L.-D. Liao, C.-T. Lin, K.~McDowell, A.~Wickenden, K.~Gramann, T.-P. Jung, L.-W.
  Ko, and J.-Y. Chang, ``Biosensor technologies for augmented brain-computer
  interfaces in the next decades,'' \emph{Proc. of the {IEEE}}, vol. 100,
  no.~2, pp. 1553--1566, 2012.

\bibitem{Lin2005d}
C.~T. Lin, R.~C. Wu, S.~F. Liang, T.~Y. Huang, W.~H. Chao, Y.~J. Chen, and
  T.~P. Jung, ``{EEG}-based drowsiness estimation for safety driving using
  independent component analysis,'' \emph{{IEEE} Trans. on Circuits and
  Systems}, vol.~52, pp. 2726--2738, 2005.

\bibitem{Lin2008}
C.-T. Lin, Y.-C. Chen, T.-Y. Huang, T.-T. Chiu, L.-W. Ko, S.-F. Liang, H.-Y.
  Hsieh, S.-H. Hsu, and J.-R. Duann, ``Development of wireless brain computer
  interface with embedded multitask scheduling and its application on real-time
  driver's drowsiness detection and warning,'' \emph{{IEEE} Trans. on
  Biomedical Engineering}, vol.~55, no.~5, pp. 1582--1591, 2008.

\bibitem{Lin2006}
C.-T. Lin, L.-W. Ko, I.-F. Chung, T.-Y. Huang, Y.-C. Chen, T.-P. Jung, and
  S.-F. Liang, ``Adaptive {EEG}-based alertness estimation system by using
  {ICA}-based fuzzy neural networks,'' \emph{{IEEE} Trans. on Circuits and
  Systems-I}, vol.~53, no.~11, pp. 2469--2476, 2006.

\bibitem{Makeig2012}
S.~Makeig, C.~Kothe, T.~Mullen, N.~Bigdely-Shamlo, Z.~Zhang, and
  K.~Kreutz-Delgado, ``Evolving signal processing for brain-computer
  interfaces,'' \emph{Proc. of the {IEEE}}, vol. 100, no. Special Centennial
  Issue, pp. 1567--1584, 2012.

\bibitem{Maynard1997}
E.~M. Maynard, C.~T. Nordhausen, and R.~A. Normann, ``The {Utah} intracortical
  electrode array: a recording structure for potential brain-computer
  interfaces,'' \emph{Electroencephalography and clinical neurophysiology},
  vol. 102, no.~3, pp. 228--239, 1997.

\bibitem{McFarland1997}
D.~J. McFarland, L.~M. McCane, S.~V. David, and J.~R. Wolpaw, ``Spatial filter
  selection for {EEG}-based communication,'' \emph{Electroencephalography and
  clinical Neurophysiology}, vol. 103, pp. 386--394, 1997.

\bibitem{Mellinger2007}
J.~Mellinger, G.~Schalk, C.~Braun, H.~Preissl, W.~Rosenstiel, N.~Birbaumer, and
  A.~Kubler, ``An {MEG}-based brain-computer interface ({BCI}),''
  \emph{Neuroimage}, vol.~36, no.~3, pp. 581--593, 2007.

\bibitem{Naseer2015}
N.~Naseer and K.-S. Hong, ``{fNIRS}-based brain-computer interfaces: a
  review,'' \emph{Frontiers in human neuroscience}, vol.~9, p.~3, 2015.

\bibitem{Nicolas-Alonso2012}
L.~F. Nicolas-Alonso and J.~Gomez-Gil, ``Brain computer interfaces, a review,''
  \emph{Sensors}, vol.~12, no.~2, pp. 1211--1279, 2012.

\bibitem{Pei2011}
X.~Pei, D.~L. Barbour, E.~C. Leuthardt, and G.~Schalk, ``Decoding vowels and
  consonants in spoken and imagined words using electrocorticographic signals
  in humans,'' \emph{Journal of neural engineering}, vol.~8, no.~4, 2011.

\bibitem{Ragin2000}
C.~C. Ragin, \emph{Fuzzy-set social science}.\hskip 1em plus 0.5em minus
  0.4em\relax Chicago, IL: The University of Chicago Press, 2000.

\bibitem{Ramoser2000}
H.~Ramoser, J.~Muller-Gerking, and G.~Pfurtscheller, ``Optimal spatial
  filtering of single trial {EEG} during imagined hand movement,'' \emph{{IEEE}
  Trans. on Rehabilitation Engineering}, vol.~8, no.~4, pp. 441--446, 2000.

\bibitem{Rivet2009}
B.~Rivet, A.~Souloumiac, V.~Attina, and G.~Gibert, ``{xDAWN} algorithm to
  enhance evoked potentials: application to brain-computer interface,''
  \emph{{IEEE} Trans. on Biomedical Engineering}, vol.~56, no.~8, pp.
  2035--2043, 2009.

\bibitem{Rivet2011}
B.~Rivet, H.~Cecotti, A.~Souloumiac, E.~Maby, and J.~Mattout, ``Theoretical
  analysis of {xDAWN} algorithm: application to an efficient sensor selection
  in a {P300} {BCI},'' in \emph{Proc. 19th European Signal Processing
  Conference}, Barcelona, Spain, August 2011, pp. 1382--1386.

\bibitem{Rivet2013}
B.~Rivet and A.~Souloumiac, ``Optimal linear spatial filters for event-related
  potentials based on a spatio-temporal model: Asymptotical performance
  analysis,'' \emph{Signal Processing}, vol.~93, no.~2, pp. 387--398, 2013.

\bibitem{Roy2015}
R.~N. Roy, S.~Bonnet, S.~Charbonnier, P.~Jallon, and A.~Campagne, ``A
  comparison of {ERP} spatial filtering methods for optimal mental workload
  estimation,'' in \emph{Proc. 37th Annual Int'l Conf. of the {IEEE}
  Engineering in Medicine and Biology Society ({EMBC})}, 2015, pp. 7254--7257.

\bibitem{Russell2015}
\BIBentryALTinterwordspacing
C.~Russell, J.~Caldwell, D.~Arand, L.~Myers, P.~Wubbels, and H.~Downs. (2015)
  Validation of the fatigue science readiband actigraph and associated
  sleep/wake classification algorithms. Accessed: 08/11/2016. [Online].
  Available:
  \url{http://static1.squarespace.com/static/550af02ae4b0cf85628d981a/t/5526c99ee4b019412c323758/14286053423
  03/Readiband_Validation.pdf.}
\BIBentrySTDinterwordspacing

\bibitem{Sagberg2004}
F.~Sagberg, P.~Jackson, H.-P. Kruger, A.~Muzer, and A.~Williams, ``Fatigue,
  sleepiness and reduced alertness as risk factors in driving,'' Institute of
  Transport Economics, Oslo, Tech. Rep. TOI Report 739/2004, 2004.

\bibitem{Sitaram2007}
R.~Sitaram, A.~Caria, R.~Veit, T.~Gaber, G.~Rota, A.~Kuebler, and N.~Birbaumer,
  ``{fMRI} brain-computer interface: a tool for neuroscientific research and
  treatment,'' \emph{Computational intelligence and neuroscience}, 2007.

\bibitem{Spuler2014}
M.~Spuler, A.~Walter, W.~Rosenstiel, and M.~Bogdan, ``Spatial filtering based
  on canonical correlation analysis for classification of evoked or
  event-related potentials in {EEG} data,'' \emph{{IEEE} Trans. on Neural
  Systems and Rehabilitation Engineering}, vol.~22, no.~6, pp. 1097--1103,
  2014.

\bibitem{Stikic2011}
M.~Stikic, R.~R. Johnson, D.~J. Levendowski, D.~P. Popovic, R.~E. Olmstead, and
  C.~Berka, ``{EEG}-derived estimators of present and future cognitive
  performance,'' \emph{Frontiers in Human Neuroscience}, vol.~5, 2011.

\bibitem{Tan2010}
D.~S. Tan and A.~Nijholt, Eds., \emph{Brain-Computer Interfaces: Applying our
  Minds to Human-Computer Interaction}.\hskip 1em plus 0.5em minus 0.4em\relax
  London: Springer, 2010.

\bibitem{Teplan2002}
M.~Teplan, ``Fundamentals of {EEG} measurement,'' \emph{Measurement Science
  Review}, vol.~2, no.~2, pp. 1--11, 2002.

\bibitem{Uriguen2015}
J.~A. Uriguen and B.~Garcia-Zapirain, ``{EEG} artifact removal --
  state-of-the-art and guidelines,'' \emph{Journal of Neural Engineering},
  vol.~12, no.~3, 2015.

\bibitem{USDoD}
{US Department of Defense Office of the Secretary of Defense}, ``Code of
  federal regulations protection of human subjects,'' \emph{Government Printing
  Office}, no. 32 CFR 19, 1999.

\bibitem{USArmy}
{US Department of the Army}, ``Use of volunteers as subjects of research,''
  \emph{Government Printing Office}, no. AR 70-25, 1990.

\bibitem{NHTSA2011}
\BIBentryALTinterwordspacing
(2011) Traffic safety facts crash stats: drowsy driving. US Department of
  Transportation, National Highway Traffic Safety Administration. Washington,
  DC. [Online]. Available: \url{http://www-nrd.nhtsa.dot.gov/pubs/811449.pdf}
\BIBentrySTDinterwordspacing

\bibitem{Erp2012}
J.~van Erp, F.~Lotte, and M.~Tangermann, ``Brain-computer interfaces: Beyond
  medical applications,'' \emph{Computer}, vol.~45, no.~4, pp. 26--34, 2012.

\bibitem{Vigario2000}
R.~Vigario, J.~Sarela, V.~Jousmiki, M.~Hamalainen, and E.~Oja, ``Independent
  component approach to the analysis of {EEG} and {MEG} recordings,''
  \emph{{IEEE} Trans. on Biomedical Engineering}, vol.~47, no.~5, pp. 589--593,
  2000.

\bibitem{Wang1997}
L.-X. Wang, \emph{A Course in Fuzzy Systems and Control}.\hskip 1em plus 0.5em
  minus 0.4em\relax Upper Saddle River, NJ: Prentice Hall, 1997.

\bibitem{Wei2015}
C.-S. Wei, Y.-P. Lin, Y.-T. Wang, T.-P. Jung, N.~Bigdely-Shamlo, and C.-T. Lin,
  ``Selective transfer learning for {EEG}-based drowsiness detection,'' in
  \emph{Proc. {IEEE} Int'l Conf. on Systems, Man and Cybernetics}, Hong Kong,
  October 2015.

\bibitem{Welch1967}
P.~Welch, ``The use of fast {F}ourier transform for the estimation of power
  spectra: A method based on time averaging over short, modified
  periodograms,'' \emph{{IEEE} Trans. on Audio Electroacoustics}, vol.~15, pp.
  70--73, 1967.

\bibitem{drwuTHMS2016}
D.~Wu, ``Online and offline domain adaptation for reducing {BCI} calibration
  effort,'' \emph{{IEEE} Trans. on Human-Machine Systems}, 2016, in press.

\bibitem{drwuaBCI2015}
D.~Wu, C.-H. Chuang, and C.-T. Lin, ``Online driver's drowsiness estimation
  using domain adaptation with model fusion,'' in \emph{Proc. Int'l Conf. on
  Affective Computing and Intelligent Interaction}, Xi'an, China, September
  2015, pp. 904--910.

\bibitem{drwuEBMAL2016}
D.~Wu, V.~J. Lawhern, S.~Gordon, B.~J. Lance, and C.-T. Lin, ``Offline
  {EEG}-based driver drowsiness estimation using enhanced batch-mode active
  learning ({EBMAL}) for regression,'' in \emph{Proc. {IEEE} Int'l Conf. on
  Systems, Man and Cybernetics}, Budapest, Hungary, October 2016.

\bibitem{drwuSMLR2016}
D.~Wu, V.~J. Lawhern, S.~Gordon, B.~J. Lance, and C.-T. Lin, ``Spectral meta-learner for regression {(SMLR)} model aggregation:
  Towards calibrationless brain-computer interface ({BCI}),'' in \emph{Proc.
  {IEEE} Int'l Conf. on Systems, Man and Cybernetics}, Budapest, Hungary,
  October 2016.

\bibitem{drwuTFS2016}
D.~Wu, V.~J. Lawhern, S.~Gordon, B.~J. Lance, and C.-T. Lin, ``Driver drowsiness estimation from {EEG} signals using online weighted
  adaptation regularization for regression ({OwARR}),'' \emph{{IEEE} Trans. on
  Fuzzy Systems}, 2016, in press.

\bibitem{Zadeh1965}
L.~A. Zadeh, ``Fuzzy sets,'' \emph{Information and Control}, vol.~8, pp.
  338--353, 1965.

\bibitem{Zhu2004}
X.~Zhu and X.~Wu, ``Class noise vs. attribute noise: A quantitative study of
  their impacts,'' \emph{Artificial Intelligence Review}, vol.~22, pp.
  177--210, 2004.

\end{thebibliography}
\end{document}